# Time Dependency, Data Flow, and Competitive Advantage

Ehsan Valavi
Joel Hestness
Marco Iansiti
Newsha Ardalani
Feng Zhu
Karim R. Lakhani

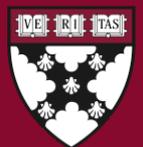

# Time Dependency, Data Flow, and Competitive Advantage

Ehsan Valavi
Harvard Business School

Newsha Ardalani
Facebook AI Research

Joel Hestness
Cerebras Systems

Feng Zhu
Harvard Business School

Marco Iansiti
Harvard Business School

Karim R. Lakhani
Harvard Business School

Funding for this research was provided in part by Harvard Business School.

# Time Dependency, Data Flow, and Competitive Advantage

Ehsan Valavi[1,2], Joel Hestness[3], Marco Iansiti[1,2], Newsha Ardalani[4], Feng Zhu[1,2] and Karim R. Lakhani[1,2]

evalavi@hbs.edu, joel@cerebras.net, miansiti@hbs.edu, new@fb.com, fzhu@hbs.edu, klakhani@hbs.edu

Harvard Business School, Boston, Massachusetts.[1]
Laboratory for Innovation Science at Harvard, Alston, Massachusetts.[2]
Cerebras Systems, Los Altos, California.[3]
Facebook AI Research, Palo Alto, California.[4]

## Summary:

The dramatic adoption of data-driven applications across different economic sectors calls for a deeper investigation into the value of data as a strategic asset[1,2,3,4]. Data is fundamental to machine learning-based products and services and is considered strategic due to its externalities for businesses, governments, non-profits, and more generally for society[5,6]. These externalities generate new forces within an industry that alter competitive dynamics, and consequently, call for new strategies, management practices and have significant implications for policymakers and regulators[7,8,9,10,11,12,13].

It is renowned that the value of organizations (businesses, government agencies and programs, and even industries) scales with the volume of available data. What is often less appreciated is that the data value in making useful organizational predictions will range widely. How the value of organizations scales with data volume is prominently a function of data characteristics and underlying algorithms.

One of the essential data characteristics is its time-dependency[14]. Time-dependency is a frequent attribute resulting from many factors, such as changes in personal preferences, location, or changes in market, industry, or societal trends. The extent of time-dependency varies across business areas and applications. Hence, understating time-dependency's extent and influence is salient not only to businesses and organizations that will rely on time-dependent data but also to regulators now faced with designing policy to guard against bias, enhance user privacy, increase consumer welfare and safeguard competition.

In this research, our goal is to study how the value of data changes over time and how this change varies across contexts and business areas (e.g. next word prediction in the context of history, sports, politics). We focus on data from Reddit.com and compare the time-dependency across various Reddit topics (Subreddits). We make this comparison by measuring the rate at which user-generated text data loses its relevance to the algorithmic prediction of conversations. We show that different subreddits have different rates of relevance decline over time. The decay rate on slow-varying subreddit like "history" is very low, as data maintains its value indefinitely. In contrast, subreddits like "world news" have a significantly higher decay rate and lose their value relatively quickly.

Relating the text topics to various business areas of interest, we argue that competing in a business area in which data value decays rapidly alters strategies to acquire competitive advantage[14]. When data value decays rapidly, access to a continuous *flow of data* will be more valuable than access to a fixed *stock* of data. In this kind of setting, improving user engagement and increasing user-base help creating and maintaining a competitive advantage.[15]

# Main Text

Virtually, for every industry, data-driven externalities[6] create forces that shape the way businesses compete. Notably, data availability can create a growth cycle between data volume and algorithmic quality[16,17]: more data leads to a better quality of products and services, which in turn increases demand[18]; the increase in demand leads to an even higher volume of data and thereby completes the cycle.

The magnitude of these competitive forces is subject to change and depends on data characteristics, such as time-dependency[14]. Time-dependency refers to the attribute that data's relevance and merit in making accurate predictions decline over time. In other words, data often is a non-durable asset, and its value perishes over time. For example, the advertisement data that is valuable now in predicting a person's purchasing preferences may be much less valuable tomorrow, next week, or next year, as the person's preferences will change. This kind of time-dependency can dramatically alter the balance of competitive advantage and transform data's influence in creating "moats," barriers to the entry of a new competitor.

Data is time-dependent for many reasons, such as changes in consumers' taste or behavior, environmental and contextual factors, or even innovation in products and services. Innovation plays a crucial role. It creates new needs that are hard to predict from old data and makes older solutions and technologies less relevant. Since the speed of change and innovation is different between contexts, the rate at which data loses its value should vary from context to context.

This research, using natural language processing models and naturally occurring consumers' text data from Reddit.com[19], shows unequal time-dependency and speed of change among different text topics representing various interest areas. We measure the change in data value for different subreddits and show they perish with different rates. For example, the value of data in the "relationship" subreddit, perishes much slowly than the value of data from "world news" and "politics" subreddits.

Reddit is a social news aggregation platform founded in 2005. As of January 2021, according to Alexa internet[A], Reddit is the 18th most visited website worldwide and 7th in the United States. 49% of the traffic is from the U.S. following by 10% and 5% from the United Kingdom and Canada. It has around 330 million monthly active users. On Reddit, users share their opinions on many different issues and contribute to multiple discussions.

Similar to Valavi et al. [14], we train a small variant of GPT-2, the Generative Pre-Training transformer-based model from OpenAI [20,21], for the next word prediction task. The next word prediction algorithm predicts the next word in a sentence given a sequence of words. We use cross-entropy as our loss function and choose the dataset size of 100MB that allows us to stay in the learning curve's power-law region[18]. We measure the effectiveness curve[14] and fit an exponential function to the measurements.

---

[A] https://www.alexa.com/siteinfo/reddit.com

We train the algorithm with large dataset sizes to assure it is adequately tuned to linguistic models. Consequently, at this level of training, we believe errors in predicting the next words mostly stem from the changes in different topics. For example, in the computer operating system topic space, after the sequence, "download windows," we may expect 'XP' in 2002 and '10' in 2020. Similarly, in science, as time goes by, researchers propose better experimentation methods and may find altered results. For example, if they claimed in 2002 that "Coffee drinking is good for heart disease" and then change the claim in 2020 to "Coffee drinking is not good for heart disease," the next word prediction algorithm picks up this development. As a result, much like the vast literature on online word-of-mouth and its economic implications,[22,23,24,25,26,27] we believe that our findings in this research are informative about the speed of change and innovation in various business areas.

**Perishability Measurement Method**

Data perishability studies change in the value of data over time. We define a metric called data effectiveness[14] to capture data's relative effectiveness in making predictions at every point in time. The perishability is then to see how the data effectiveness changes over time.

We elaborate on how to measure data effectiveness using an example shown in Figure 1. As depicted in Figure 1(i), we train a model on the dataset ($A$) of size $|A|$ sampled from time 0. We then evaluate the model's performance (Loss value) on a testing dataset sampled from time $T$. Let's say the model produces the loss value $L$.

We then use a training dataset sampled at time $T$ to see what training set size from this sample (if tested on a dataset from time $T$) would result in a similar loss value. Let's say size $|B|$ from time $T$ reaches $L$. $|B|$ is expected to be less than $|A|$ since $A$ has lost its predictive relevance over time. We define the ratio $\frac{|B|}{|A|} \in [0,1]$ as dataset $A$'s effectiveness at time $T$.

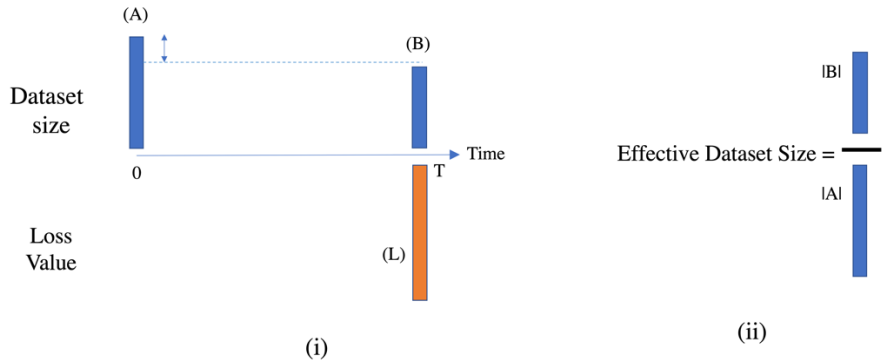

*Figure 1) These figures conceptually illustrate the loss in data value due to time-dependency. The bars below the time axis represent the loss value, and the bars above the time axis are dataset sizes. In figure 1 (i), (A) represents a training set with size |A| that is sampled at time 0. (L) is the loss value measured for the model trained on the dataset (A) and tested at time T. (B) represent a training dataset with size |B| from time T that if we train the model using (B) and test it at T, the loss value would be (L). Figure 1 (ii) shows the dataset's effectiveness value. The perishability curve is then to measure effective datasets sizes for multiple T and study the evolution of effectiveness.*

**Perishability Curves Track Real-World Changes**

To reassure that our method tracks real-world changes, we look into a few subreddits' perishability curves. For example, in Figure 2, we measure the perishability of datasets from October 2012 and see if the measured changes correspond with real-world events. We can see periodicity in perishability of sports datasets like "hockey" (This is expected since we have seasonality in sport) or a flat perishability curve in "history" (This is expected since commentators in such forums usually discuss events far in the past). Yet, the most interesting behavior arises in the "politics".

Figure 3 presents the perishability curves of the "politics" subreddit. We observe that the value of 2012's data declined mostly in mid 2015. The observation indicates that political discussions in 2015 and 2016 (Before 2016's presidential election) are not predictable from 2012's data, and we suspect a drastic shift in the political landscape and a change in political discussions on those years. These observation highlights our method's functionality in tracking changes over time.

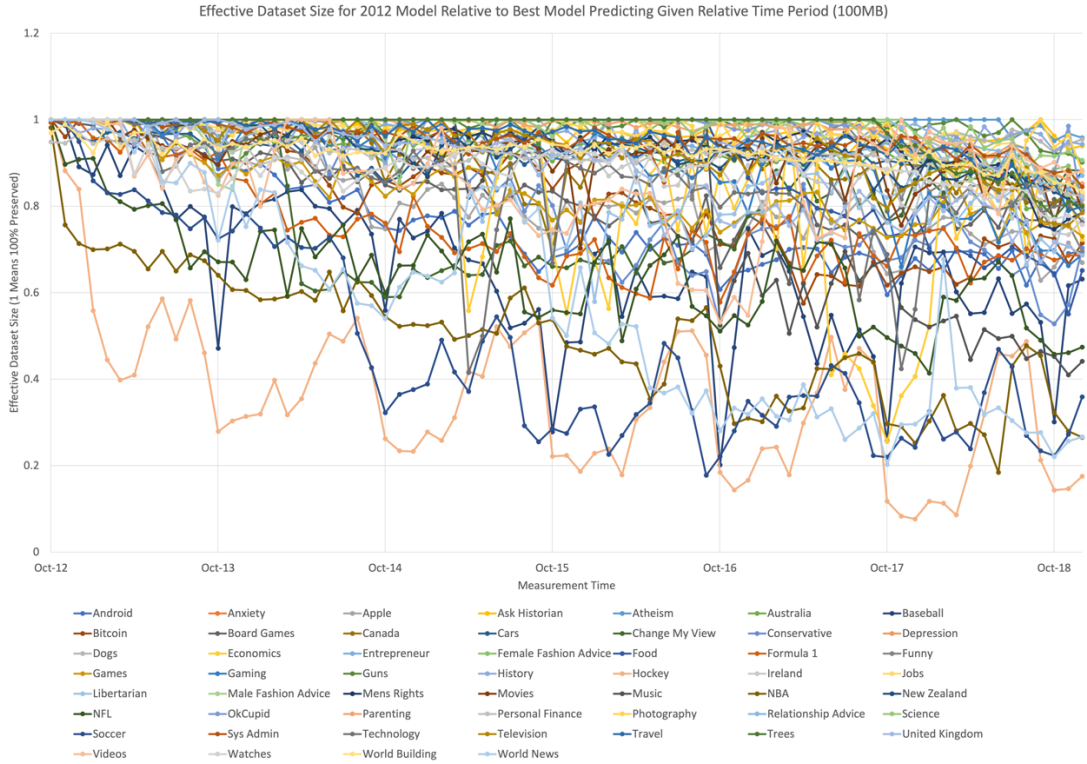

*Figure 2) Effective dataset sizes for multiple topics. This graph visually shows the difference in data perishability between the topics. It is worth noting that for most subreddits, the dataset's effectiveness is not constantly diminishing. We see ripples or sudden drops in value. The general trend, though, shows a decrease in the overall value of data over time.*

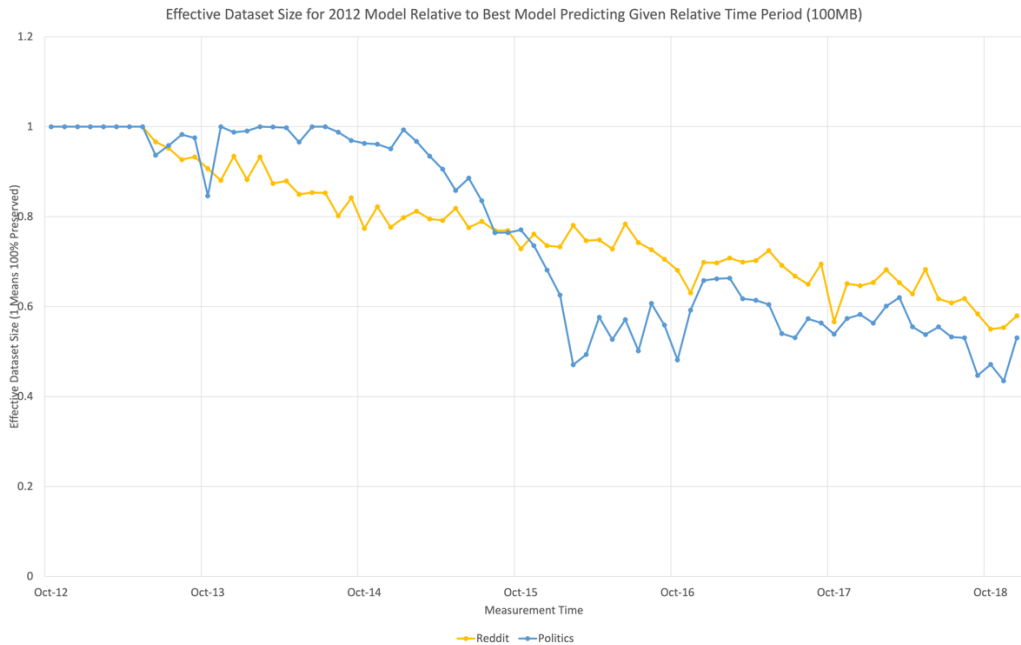

*Figure 3) Effective dataset sizes for the politics subreddit (Blue) and entire Reddit data (Yellow). The dataset from all Reddit topics has a monotonic decrease in value. In contrast, the dataset sampled in October 2012 from the politics subreddit perishes mainly at the beginning of 2015 and reaches the lowest point in February 2016 (The months leading to the 2016 United States presidential election)*

**Characterizing the Perishability Trends**

The perishability curves in Figure 2 don't lend themselves to a unique functional form. We can observe a macro trend for each curve, showing an overall decline in the value of data and a micro-trend that often manifests itself with periodicity. The micro-trend is particularly visible in the "hockey" subreddit dataset in Figure 2. For the rest of this section, we focus on the macro trends and characterize the overall decline rate in the data value for the entire Reddit dataset and a few subreddits.

We find (Explained in supplementary information section) that exponential function fits the decay trend best. Table 1 shows the estimated exponential decay rates $\mu$ from the functional form $e^{-\mu t}$ for different subreddits. It is estimated using

$$\log(y) = -\mu t + u \qquad (1)$$

where $y$ is the measured effectiveness[14], $t$ is time (in years), and $u$ is normally distributed fitting noise. Since decay rates might be hard to interpret, we also provided the dataset half-life-time. Half-life-time is the period that it takes for a dataset to loses half of its predictive substance. , i.e., the time $t$ where $e^{-\mu t} = \frac{1}{2}$. Figure 4 provides the half-life-time for multiple subreddit datasets.

*Table 1) Perishability rate measurements for several topics.*

| | ESTIMATE ($-\mu \sim \frac{1}{year}$) | HALF-LIFE-TIME (YEARS) | STANDARD ERROR (ALL ESTIMATES ARE SIGNIFICANT AT $10^{-3}$) |
|---|---|---|---|
| HISTORY | -0.004 | 100> (168.9) | 6.56E-04 |
| RELATIONSHIP | -0.010 | 66.69 | 4.71E-04 |
| MOVIES | -0.026 | 26.53 | 0.001 |
| FOOD | -0.048 | 14.39 | 0.002 |
| TECHNOLOGY | -0.054 | 12.78 | 0.003 |
| APPLE (THE COMPANY) | -0.059 | 11.76 | 0.001 |
| ENTIRE REDDIT | -0.084 | 8.22 | 0.001 |
| NFL | -0.100 | 6.92 | 0.003 |
| MUSIC | -0.108 | 6.43 | 0.004 |
| BASEBALL | -0.122 | 5.67 | 0.005 |
| POLITICS | -0.151 | 4.58 | 0.004 |
| NBA | -0.189 | 3.67 | 0.004 |
| SOCCER | -0.230 | 3.00 | 0.006 |
| WORLD NEWS | -0.233 | 2.97 | 0.005 |
| HOCKEY | -0.245 | 2.83 | 0.009 |

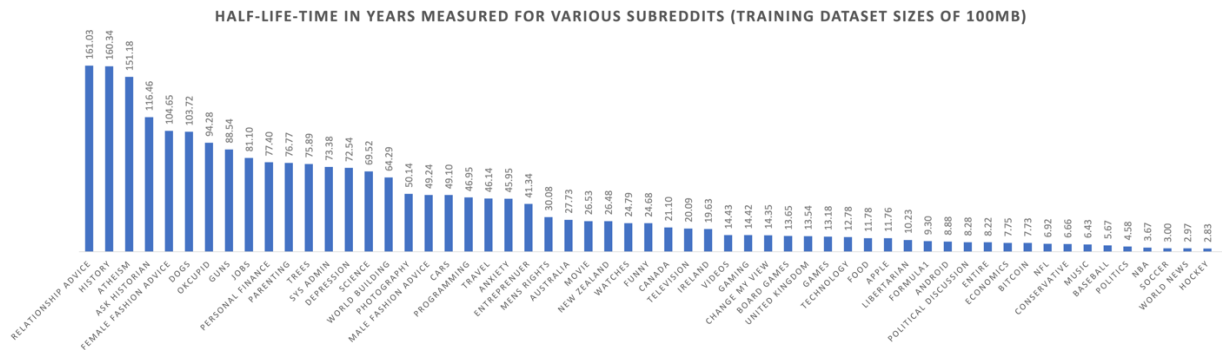

*Figure 4) Half-life-time measured for a few most visited subreddits. We see sports, politics and economics on the right side (Higher perishability) followed by technology, society and leisure in the middle. On the far left, we see topics mostly related to personal development.*

It is interesting to see that subreddits like "relationship" appearing to be durable. In other words, the way people talk about relationships is not changing quickly. In contrast, it takes 4-5 years for a dataset of 100MB in "politics" to lose half of its predictive substance.

Two other interesting topics arise from United States professional sports setting by comparing the National Basketball Association (NBA) and the National Football League (NFL). They have significantly different half-life-time. We believe it is due to their structural differences. To name one, the number of games each team plays each season in the NBA is 82, whereas it is 16 games in the NFL. In total, the NFL has 256 games per season, which is much smaller than the NBA. Therefore, the opportunity of new events in the NBA is naturally higher, which accounts for 3.67 years of half-life time comparing to 6.92 years in NFL. Similarly, National Hockey League schedules 82 games per team, and therefore the subreddit "hockey" has perishability rate similar to NBA's.

Other factors like players' longevity and movements are essential in the predictability of events. Basketball, football, and hockey are physically demanding sports and have roughly similar players' longevity. Hence, the number of games is a good proxy for measuring the number of new events. In contrast, the baseball subreddit, despite MLB's 162 games per season, has lower perishability. It means that the rate of new events per season is lower for MLB comparing with NFL, NBA, or NHL. A good explanation for this is players' longevity, which is higher in baseball.

**Pairwise Comparison of Macro Trends**

In the previous section, we measured the decay rates for a few subreddits and observed varying degrees of perishability over different topics. This section wants to certify that the decay rates (macro trends) are different between subreddits, i.e., to see if two datasets have indistinguishable macro trends if they have similar perishability rates. To answer this question, we formulate a new test that captures the difference between every subreddit pairs as follows:

$$\log(y_i) - \log(y_j) = -\beta t + \epsilon \quad (2)$$

Where $y_i$ is the measured effectiveness for topic $i \in \{relationship, history, ...\}$, $t$ is time, and $\epsilon$ is the normally distributed noise. The coefficient $\beta$ should be zero if two topics have identical decay rate. Therefore, in comparing different topics, the null hypothesis would be $\beta = 0$.

Table 2 shows the p-value for subreddit pair's $\beta$ estimates. We expect a higher p-value (shown in darker blue) if two datasets have relatively indistinguishable macro trends. Statistically speaking, darker blue means that a significant difference in perishability rates is not evident from our data using the exponential decay model.

In Table 2, we see about ten dark blue clusters of subreddit pairs. Though we can't establish a causal relationship between the subreddits in each cluster, it is still worth noting that a casual relation would lead pairs of subreddits to belong to the same cluster. For example, we expect subreddits like "gaming", "board games," and "games" to belong to the same cluster. We don't want to draw a causal conclusion, but it is interesting to see that the pairs (economics, conservative) and (economics, bitcoin) have dark blue in table 2. Yet, the pair (conservative, bitcoin) has a light blue color.

*Table 2) P-Values for estimated β (The difference between perishability rates). There are four colors in this graph. Very light blue means that the subreddit pair's estimated β is different from zero at 10-3 significance level. Light blue and blue colors show significance at 0.01 and 0.05 levels, respectively. The darkest blue color indicates that the p-value is higher than 0.05, meaning that for the exponential decay model, a significant difference in perishability rates is not evident from our data.*

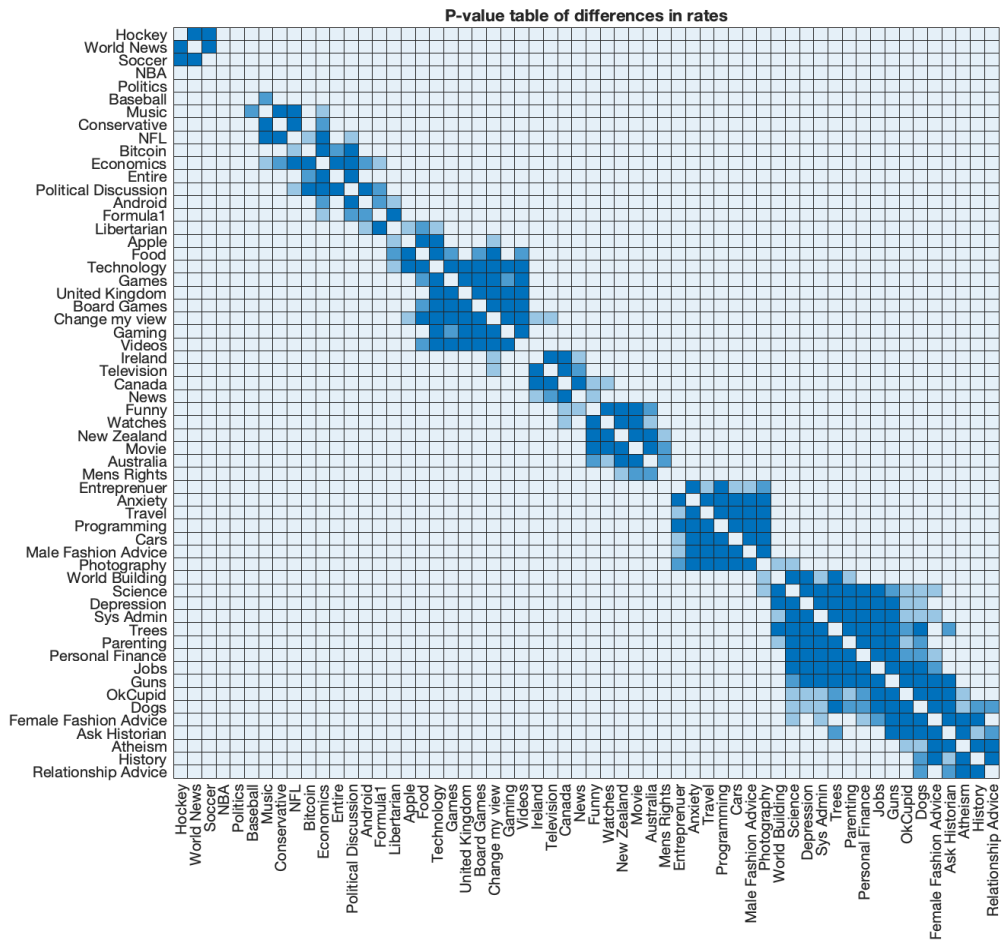

**Implications**

Data perishability has strategic implications for businesses that provide data-driven products and services. High perishability undermines the importance of data volume or historical data in creating a competitive advantage[14]. On the other hand, as proved in the supplementary information section and also suggested by Claussen et al.[28], increasing the flow of data can compensate for the volume's value loss due to perishability.

Our findings directly influence practices benefiting from user-generated text data[29,30,31,32,33]. For example, services and products like search engines, recommendation systems, and AI-enabled personal assistants and translators can adjust the algorithms and training data repository to account for the importance of data flow in different contexts.

Although our perishability analysis has been limited to products and businesses driven by the language models, we believe the conclusion that the data perishability rate is a function of dataset can be extended to other businesses. In other words, our method indirectly illustrates how different business areas might differ with respect to the rate of decay in data value and hence, the importance of data flow in their operations. Subsequently, business areas facing higher perishability offload[14] historical data more frequently which not only reduces the dataset size and the complexity of operation, but it increases the effectiveness of the datasets. Besides, removing unnecessary data in the offloading process answers some of the users' privacy concerns.

To increase the data flow, businesses can, for example, increase user engagement or expand the user-base when there exist user-wide externalities. The importance of growing user-base highlights the possibility of market dominance and concentration[34], making our findings relevant to antitrust debates. Therefore, from the regulatory perspective, authorities should be aware of the profound difference in the value creation dynamics across business areas. They should craft policies considering economic models that include the flow dynamics for business areas facing higher perishability than models solely based on data stock.

Our method in measuring a dataset's effectiveness can be used in other research areas such as linguistic, economics, and social sciences. For example, one can study the socio-economic impact of innovation, policy introduction, or stimuli on the user's behavior by observing user-generated text data's predictability. While in our paper, we used the next word prediction task to track behavior changes, other types of prediction tasks can be used depending on the research domain and question.

The supplementary information section contains proofs, details of implementation, and a broader discussion on the choices we made in this paper.

# Supplementary Information Section

1. **Why Time-Dependency?**

Data is time-dependent for many reasons, among which we can mention the change in consumers' taste or behavior and, more importantly, innovation in products and services. Innovation plays a key role in time-dependency. It is because it creates new needs that are hard to predict from old data. Besides, due to innovation, older solutions and technologies are gradually being eliminated, which means that data becomes irrelevant in some cases.

As an example, consider data on kerosene's price and consumption. Before electricity, kerosene was used on lamps to light homes and offices. With the invention of electric bulbs, kerosene lost its lighting purpose, and rare is a time to see it being used for lighting. Consequently, consumers have different price elasticity, which means using kerosene's price data to study macroeconomic questions may not be as relevant as it was many years ago. Besides, the invention of electricity created new consumer needs like refrigerators, air conditioning systems, the Internet, and social media, which changes consumers' behavior.

In kerosene's example, it took decades to witness the change, and the speed of decline for data appeared to be slow. In contrast, in electronics and particularly the cellphone business, change happens at a faster pace. Less than two decades ago, smartphones were introduced, and with them came many innovations in communication methods and devices. Because of these changes, earlier cellphones are becoming less usable and, in some cases, not even compatible with telecom infrastructure.

Similar to our argument on kerosene, using data on old cellphones is not as relevant as using recent cellphone's data. It means the speed of change in the cellphone business is even faster than the speed of change in the energy sector. These observations are market-specific and are affecting every firm within a market. For example, if consumers' taste changes fast, all firms should follow the change quickly or lose the business. Because of it, the speed of change has market level consequences and may change modes of competition.

## 2. Advantages over Alternative Measurement Methods:

One way to compare different markets/industries with respect to their speed of change is to create a pool of companies from different industries and study how relevant are the old data to their current problems. For example, we can take companies like Uber in the rideshare business and New York Times from news and media to study how consumer's data lose value in time for those specific companies and generalize it to their industry, respectively.

This method has several challenges. To name a few, we can mention selection biases, algorithmic differences, and availability. About selection biases, not everyone is doing business with Uber, and similarly, New York Times readers have a particular taste. Therefore, there are biases in how data is generated, making the comparison difficult, and not generalizable to other companies in the same industry.

As of algorithmic difference, we can immediately tell that New York Times and Uber are in different businesses, and therefore, they need data for different purposes. Besides, perishability changes with the learning curve and scalability of algorithms. Since NYT and Uber use user's data for different tasks and use different algorithms with distinct scaling behavior, perishability measures, as defined in Valavi et al.[14] are not directly comparable.

Finally, even if we solve selection biases and algorithmic differences issues, it is not easy, if not impossible, to get users data from companies to do the analysis.

## 3. Design of Experiment:

Data Collection and Processing

We choose the Reddit post dataset as it fits our needs (Data was collected for Fan et al.[19]). This dataset is a collection of posts and comments from 2006 to 2018. It was scraped from Reddit between September 2006 and July 2018. We preprocessed the dataset to create flat text files with the following form:

```
Title (6): What was the biggest scandal in your school?
Text:
Comment (4): Vampires. This was almost 6 years ago now at my
high school, but vampires. Do a quick...
```

```
Comment (3): Not sure if I'd call it a "scandal," but when I
was in college...
Comment (2): Freshman year a friend of mine found a paper
bag at the bus stop full of money - and it...
```

The 'Title' is the title that the author specified when posting the submission, and 'Text' is an optional field of body text associated with the post. After the post, each line is a comment from other users designated by 'Comment'. The comments only contain text. The values in parenthesis are submission or comment scores based on upvotes or downvotes given to each by users. We filtered posts and comments with scores less than 2.

We split the dataset into chunks based on the submissions and comments' timestamp. We aim for 100 million words per time period, so we group data together until each split is at least that large. Specifically, we group posts and comments into monthly periods for the years 2012 to 2018 since those periods had sufficient data to investigate different topics.

Finally, we subdivide the data from each time period to form a standard machine learning training and testing setup for collecting learning curves. First, we randomly sample and split the posts (and their respective comments) into training, development/validation, and test/evaluation subsets. The development and test sets are at least 2 million words each. The development set is used to validate that the model is learning to generalize during training and to early-stop training when the model performs the best on the development set. The test set is used after training to evaluate how well the training, and we use these test sets to cross-evaluate models trained on data from other time periods. The model never trains on these subsets.

After splitting out the development and test sets, the remaining data is randomly shuffled as the full training set for the time period. We subdivide this training set into chunks of exponentially increasing size by factors of 2. Empirically, we find that datasets of size 1.25 million words are large enough to be in the power-law portion of learning curves[18], so we break the training set into successively overlapping subsets of size 40 million, 20 million, 10 million, 5 million, 2.5 million, 1.25 million words by taking the first half of the prior subset. We train separate models on each training subset to collect how models generalize as they are allowed to train with increasing dataset size. The resulting data size-generalizability curves are learning curves for the time period.

Model Architecture and Training Process

We chose to train current state-of-the-art language models on the data to collect their generalization error and learning curves. Specifically, we train GPT-2, the Generative Pre-Training transformer-based model from OpenAI[20,21]. Collecting learning curves can be costly due to the training time required to train large models on each of the training subsets. We chose to train a small variant of GPT-2 that was expected to be large enough (i.e., sufficient parameters) to overfit all of the training set splits and yet small enough to train in a reasonable amount of time---at most about 32 hours per training subset on a single GPU. We configure our GPT-2 model variant as follows: Vocabulary size 50257 sub-word tokens, maximum sequence length 512 tokens, depth 6 transformer blocks each with 8 self-attention heads, and hidden dimension 512. The model has 44.9 million parameters total—a rule of thumb in language modeling is to use a model with as many parameters as words in the largest dataset.

We train the models using the Adam optimizer with a static learning rate of 2e-4 and with batch sizes 12 and 24. The training objective is the cross-entropy loss of the model's prediction of the probability of the target next token in the input sentence. We find empirically that changing the batch size marginally changes the final loss (<0.3% change in cross-entropy), so we do not further explore optimization hyperparameters as a way to mitigate total training time. Finally, we validate the models using the development dataset every 50-200 training steps, depending on the dataset's size—smaller datasets require fewer training steps for the model to converge. We early-stop training when the development set loss stops improving for more than 15 validation runs.

Evaluation Process and Effective Dataset Size

Our objective is to evaluate how well a dataset for one time period can predict values for each other time period's data—a measure of how much the data distribution has changed over time. We start by finding the "best" model for each time period and each dataset size. During a training run of a given time period and given training set size, each training step updates the model's weights. We periodically validate the models on the given time period's development set and designate the model weights that achieve the best development set loss as the "best" model in each training run. When we test with multiple different batch sizes, we choose the best model of the separate training runs as the best model for the time period and training set size. This "best" model selection process mimics the way models are chosen for deployment in AI-enabled products.

Now, for a given time period, we have collected the best model for each training set size ranging from 1.25 to 40 million words, allowing us to construct learning curves across different time periods. We cross-evaluate all best models—one for each time period and training set size—by evaluating them on the test sets for all other time periods. We use these results to curve fit learning curves: Given the best models for the time period, $t_1$, and their evaluation scores for the time period, $t_2$ ($t_2$ can be equal to $t_1$), these scores show how increasing the training set size from period $t_1$ might improve prediction accuracy for time period $t_2$. We curve fit these scores with power-laws[17].

Finally, we invert these learning curves to estimate the equivalent dataset size from time period $t_1$ when predicting time period $t_0$. Start with the best model, $m_{t_1,50M}$, for time period $t_1$ trained on 50 million words, for example. Evaluate $m_{t_1,50M}$ to collect cross-entropy loss for time period $t_0$. Now use the learning curve for models trained and tested on time period $t_0$ to estimate how much training data from time period $t_0$ is required to achieve that cross-entropy loss. Suppose the inverted learning curve yields 40 million words required in time period $t_0$, then the equivalent dataset size from time period $t_1$ is 40 million words at time $t_0$, or it is effectively 80% of its time $t_1$ size.

4. Exponential or Power-law Decay Functions?

As shown in figure 2, there is no unique functional form describing the value loss. Sometimes, the wind of change blows strongly, and other times the entire world stops. An example of this can be seen in the politics topic in the years leading to the 2016 U.S. presidential election (Figure 3).

Therefore, we consider a functional that takes the difference between training and testing time as an input, and outputs the effective size. In other words, this functional is independent of the testing and the training times and only takes the difference as input. This is a valid assumption because of the measurements provided in Valavi et al.[14] on the value loss over entire Reddit data. In that paper, the equivalence graphs showed that text data's value decline is independent of sampling and testing times. In these graphs, equivalence appears as a function of the difference in testing and sampling times.

Besides, for this paper's purpose, which compares the speed of change between different topics, we need the function to have only one scalar parameter measuring the decay. This function is decreasing and bounded from below by zero, which means that a convex function could be a suitable candidate.

Consequently, we decide to test the exponential and the power-law functional forms since they both satisfy the mentioned conditions. The visual inspection, provided in Figure 5, suggests that the exponential function describes the value decay better between these functionals.

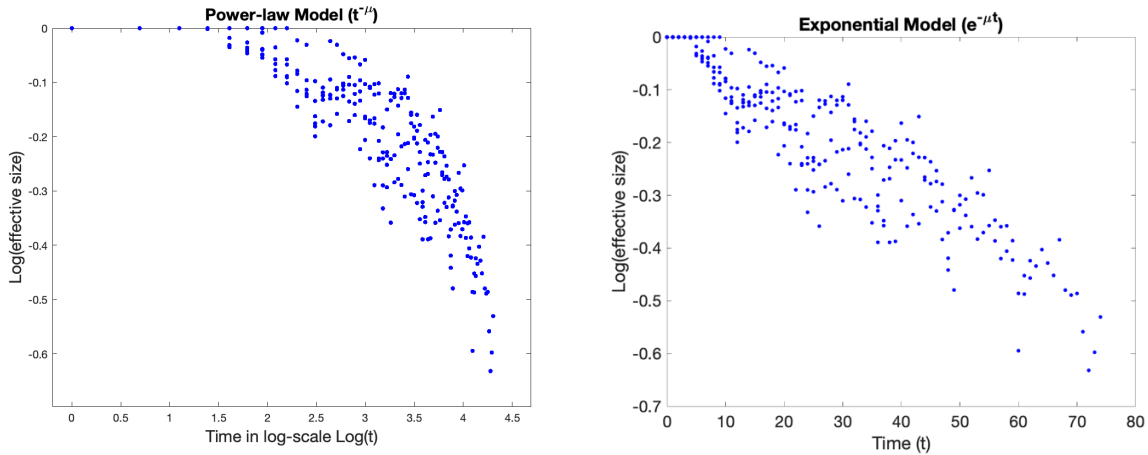

*Figure 5) Power-law (Left plot) and exponential (Right plot) curve fitting. We expect to see a linear representation in either graph.*

In Figure 5 left, we took the log from the time dimension and expected to see a linear functional form if the power-law decay model describes the measurements. As apparent from the graph, it is not linear. In contrast, taking the log from the effectiveness, it is easy to observe a linear relation meaning that the exponential decay model describes the measurements best.

### 5. Perishability and Data Flow:

As previously shown in Valavi et al.[14], for highly time-dependent businesses, the data volume does not significantly contribute to the scalability of data-driven AI solutions. To such an extent that there is a finite upper-bound on the effective size of perishable datasets. This bound shows that the data network effect cycle saturates and does not scale beyond the upper-bound even with a dataset of infinite sizes (That has shifted distribution).

In this paper, we show that data flow, to some extent, mitigates the scaling limitations. Expanding the user-base, in the presence of user-wide externalities, or increasing user engagement create the required data flow. Either way, the increase in data flow leads to an increase in the equivalent size.

The proof requires several theorems. First, we mention[14] that net distribution, in the dataset gathered over a period of time, is solely a function of the underlying data distribution and the sampling density. Hence, multiplying the dataset size by a constant that shows the increase in the flow does not change the net distribution, yet it improves the dataset's equivalent size. Second, we show that in comparison with non-perishable data, it is likelier for the perishable data to offload[14] the dataset and move the equivalent time[14] closer to the prediction time. Getting the equivalent time closer to the prediction time increases the upper-bound limit and, thereby, enhances the effectiveness of data flow in improving quality.

Putting these theorems together, we conclude that data flow derives the scale in time-dependent business. In a nutshell, the following two forces make the information flow the primary driver of value creation for a perishable dataset:
- Pushing the equivalent time closer to prediction time through off-loading, which makes historical data less critical.
- Increasing the upper-bound of equivalent size for equivalent times closer to the prediction time, which makes the flow the main scalability driver.

<u>Terminology and definitions</u>

Definition 1-4: For a dataset ($D_{n,[0,t],\lambda_t}$) of size $n$ that has been sampled since $t$ periods prior to prediction time with the density function $\lambda_t$ (Where for $\forall s \in [0,t]$, $\lambda_s \in [0,1]$ and $\int_0^t \lambda_s ds = 1$) we define following:
1) Equivalent time $t^* \in [0,t]$ is the time that an IID dataset of size $n$ from the time $t^*$ ($D_{n,t^*}$) produces the loss value equal to the loss value of $D_{n,[0,t],\lambda_t}$.

2) Equivalent size[12] $\bar{n}_{D_{n,[0,t],\lambda_t}}$ is the size (E) in figure 1. If the dataset is sampled only at time $t$ ($D_{n,t}$) we use the notation $\bar{n}_{D_{n,t}}$.

3) Effectiveness[12] $E_{D_{n,[0,t],\lambda_t}} = \frac{\bar{n}_{D_{n,[0,t],\lambda_t}}}{n}$ or alternatively $E_{D_{n,t^*}} = \frac{\bar{n}_{D_{n,t^*}}}{n}$ is the ratio of $\frac{size(E)}{size(A)}$ in figure 1.

4) Substitution function[12] $f_n(t_1, t_2) = \frac{\bar{n}_{D_{n,t_1}}}{\bar{n}_{D_{n,t_2}}}$ shows the gain in equivalent size when we substitute data from time $t_2$ with a dataset from time $t_1$.

Assumption 1:
1) The equivalent size $\bar{n}_{D_{n,t}}$ is monotonically decreasing over time.

Definition 5) Datasets from topic/business $H$ is more perishable comparing to datasets from topic/business $L$ if, fixing the sampling function $\lambda_t$ and the size for both datasets, we have

$$\bar{n}_{H_{n,t_1}} - \bar{n}_{H_{n,t_2}} > \bar{n}_{L_{n,t_1}} - \bar{n}_{L_{n,t_2}}$$

For all $t_1, t_2 \in [0, t]$ and $t_1 < t_2$.

Lemma 1) (Valavi et al.[14]) The net distribution for the dataset $D_{n,[0,t],\lambda_t}$ is equal to

$$P_{[0,t],\lambda_t}(x) = \int_0^t P_s(x) \lambda_s ds$$

Which is independent of the size. ∎

**Theorem 1)** Highly time-dependent datasets have equivalent time closer to the prediction time than less perishable datasets.

**Proof:** We focus on the off-loading mechanism and how it is reasonable for a highly perishable dataset to off-load more often than a less perishable dataset. To prove the theorem, we first prove that highly perishable data has a sharper substitution curve meaning $f_n^H(t_1, t_2) > f_n^L(t_1, t_2)$ where $H, L$ means high and low perishability. Then, we use this inequality to prove that every off-loading iteration for a low perishable dataset is also an off-loading iteration for a high perishable dataset. Therefore, we conclude that high perishable data has equivalent time closer to prediction time.

Definition 5 states that $H$ is more perishable than $L$ if

$$\bar{n}_{H_{n,t_1}} - \bar{n}_{H_{n,t_2}} > \bar{n}_{L_{n,t_1}} - \bar{n}_{L_{n,t_2}} \quad (3)$$

Since $\bar{n}_{L_{n,0}} = \bar{n}_{H_{n,0}} = n$, we alternatively have $\bar{n}_{L_{n,t}} > \bar{n}_{H_{n,t}}$ or $E_{L_{n,t}} > E_{H_{n,t}}$. In other words, data from low perishable datasets remain effective for a longer time. Therefore, we have

$$\frac{1}{\bar{n}_{H_{n,t_2}}} > \frac{1}{\bar{n}_{L_{n,t_2}}}$$

Multiplying above inequality to inequality (3) we have

$$\frac{\bar{n}_{H_{n,t_1}} - \bar{n}_{H_{n,t_2}}}{\bar{n}_{H_{n,t_2}}} > \frac{\bar{n}_{L_{n,t_1}} - \bar{n}_{L_{n,t_2}}}{\bar{n}_{L_{n,t_2}}}$$

$$\Leftrightarrow \frac{\bar{n}_{H_{n,t_1}}}{\bar{n}_{H_{n,t_2}}} - 1 > \frac{\bar{n}_{L_{n,t_1}}}{\bar{n}_{L_{n,t_2}}} - 1 \Leftrightarrow f_n^H(t_1, t_2) > f_n^L(t_1, t_2)$$

Which proves the first step. For the second step, consider a highly perishable dataset $H_{n,[0,t],\lambda_t}$ with identical sampling density function and equal size to a less perishable dataset $L_{n,[0,t]\lambda_t}$.
The condition for a successful off-loading iteration from the equivalent time $t^*$ to $t^{**}$ is[14]

$$f_{n-n_0}(t^{**}, t^*) > \frac{n}{n - n_0}$$

Since we assumed both $H_{n,[0,t],\lambda_t}$ and $L_{n,[0,t],\lambda_t}$ have identical sizes and sampling density functions, deleting identical period's data from both datasets still makes them have equal sizes. Consequently, for $t_1 < t_2$, any off-loading iteration that changes $L_{n,[0,t_2],\lambda_t}$ to $L_{n-n_0,[0,t_1],\lambda_t}$ is also an iteration for $H_{n,[0,t_2],\lambda_t}$ to $H_{n-n_0,[0,t_1],\lambda_t}$ since:

$$f_n^H(t_1, t_2) > f_n^L(t_1, t_2) \geq \frac{n}{n - n_0}$$

And that completes the proof. ∎

**Theorem 2)** (Valavi et al.[14]) The upper-bound on the equivalent size decreases in $t$. The closer the equivalent time to the prediction time, the larger the upper-bound on equivalent size.

**Proof:** The upper-bound is equal to $\bar{n}_{D_{\infty,t}}$ since $\bar{n}_{D_{n,t}}$ is increasing in the number of data points. Assuming a monotonic decline in the equivalent size means that this upper bound is indeed decreasing in time. ∎